# Revolutionizing Personalized Cancer Vaccines with NEO:

# Novel Epitope Optimization Using an Aggregated Feed Forward

# and Recurrent Neural Network with LSTM Architecture

Nishanth Basava[1], McCallie School

Computational Biology

**Abstract**

As cancer cases continue to rise, with a 2023 study from Zhejiang and Harvard predicting a 31% increase in cases and a 21% increase in deaths by 2030, the need to find more effective treatments for cancer is greater than ever before. Traditional approaches to treating cancer, such as chemotherapy, often kill healthy cells because of their lack of targetability. On the contrary, personalized cancer vaccines can utilize neoepitopes - distinctive peptides on cancer cells that are often missed by the body's immune system - that have strong binding affinities to a patient's MHC to provide a more targeted treatment approach. The selection of the perfect neoepitopes that elicit an immune response is a time-consuming and costly process because of the required inputs of modern predictive methods. This project aims to facilitate faster, cheaper, and more accurate neoepitope binding predictions with the use of Feed Forward Neural Networks and Recurrent Neural Networks.

Thus, NEO was created. NEO requires next-generation sequencing data and utilizes the stacking ensemble method by calculating scores from state-of-the-art models (MHCFlurry1.6, NetMHCstabpan-1.0, and IEDB.) The model's architecture includes an FFNN and an RNN with LSTM layers with memory to analyze both sequential and non-sequential data. Both model's results are aggregated to produce predictions. Using this model, personalized cancer vaccines can be produced with improved results (0.9166 AUC, recall = 91.67%,).

[1] **Acknowledgements:** I would like to thank Dr. Ashley Posey at the McCallie School for her mentorship and wisdom throughout this project. Additionally, Dr. Giovanni Parmigiani, Mrs. Ezhilvadivu Palaniyappan, Mrs. Neelima Maddy, and Ms. Jenna Landy are thanked for providing feedback on my project and knowledge in machine learning and artificial intelligence.

# 1 Introduction

Cancer—the notorious, pathogenic disease that is estimated to have killed almost 10 million people and caused an estimated 19.3 million new cancer cases in 2020—remains a significant global health challenge [20]. Even with improved treatment options, these values had a 26.3% and 20.9% increase from 2010 to 2019 respectively [15]. In the past, the lack of efficient sequencing techniques has limited the ability to develop personalized techniques that are tailored to a patient's tumor genome. More recently, however, with technological innovations making next-generation sequencing cheaper and more accessible, more direct patient specificity may be possible [13]. By performing next-generation sequencing on a patient's normal and tumor genome, tumor-specific antigens, commonly known as neoantigens, can be identified [3]. The issue with developing neoantigen vaccines is the difficulty in predicting whether the neoantigens would bind to the MHC-I and MHC-II of the antigen-presenting cells (APCs) and be recognized by the immune system. In vitro testing of neoantigen presentation in the immune system would be time-consuming as scientists would have to test the thousands of neoantigens present in the patient's tumor for binding to the patient's HLA. Through the usage of computational models, these predictions can be automated with a cheaper, less time-consuming, and more accurate methodology.

Over time, newer machine learning models, such as neural networks, have been developed with increased predictive power by being able to recognize patterns and optimize parameters to achieve more accurate results [14]. Within the field of neural networks, there exist different types, each of which is able to handle different data types best and learn the patterns of the training data differently. For example, feed-forward neural networks (FFNNs) typically train best on non-sequential data whereas the more novel Long-Short Term Memory (LSTM) architecture of recurrent neural networks (RNNs) is able to implement memory to learn from sequential data [22]. This project aims to utilize the power of both FFNNs and RNNs to create a two-branched, ensemble model that takes both sequential and non-sequential data as inputs and is able to produce improved predictions of neoantigen-MHC binding.

# 2 Background

## 2.1 Neoepitopes

Cancer is caused by a build of somatic mutations throughout the body cells that affect various cellular processes [1]. A key change has to do with the identity of the body's cells. On the surface of most human cells are peptides— sequences of amino acids —called epitopes. Epitopes are a part of antigens, which are substances that trigger immune responses, that can be recognized by the immune system's b-cells or t-cells as either self or non-self. If the immune cells recognize the epitope as self, central immune tolerance causes the cell to not attack itself [15]. However, if the immune cells find an epitope that it does not recognize as self, otherwise known as a neoepitope (the part of a neoantigen that binds to the MHC), then the immune system will try to get rid of the cell with the neoepitope [15]. The key to recognizing the neoepitope lies in the binding of the neoepitope to the surface of the Major Histocompatibility Complex (MHC) molecules on the antigen-presenting cell (APC). In humans, the MHC is known as the human leukocyte antigen (HLA). The APCs then mature the cytotoxic T-cells, causing a cascade of tumor killing, or immunosurveillance, that results in the release of more neoepitopes and increased binding through a positive feedback mechanism [15].

By determining which patient-specific epitopes on the tumor's surface elicit an immune response, the tumor can be accurately targeted, and improved results can be attained, especially in patients with later stages of cancer. This would result in more favorable outcomes than just targeting general cells or pathways in treatments such as chemotherapy or immunotherapy. Machine learning makes it possible to bypass running thousands of in vitro assays testing for binding affinity.

## 2.2 Proposed Method of Treatment

As mentioned earlier, traditional approaches lack targetability, so why aren't oncologists using personalized cancer vaccines? The reason lies in the inefficiency of current methodologies of personalized cancer vaccine development. After next-generation sequencing and epitope identification, computational models can be utilized to determine which of those neoepitopes are effective at eliciting an immune response. Sequencing techniques such as high throughput detection have been developed, making acquiring genomic data cheaper, faster, and easier than ever before [2]. In fact, the recent SARS-CoV-2 vaccines were able to be developed so quickly largely due to this innovation [2]. Additionally, In vitro testing of each epitope is a very time-consuming process, and through the utilization of modern machine learning and deep learning techniques such as neural networks, these lengthy assays can be bypassed.

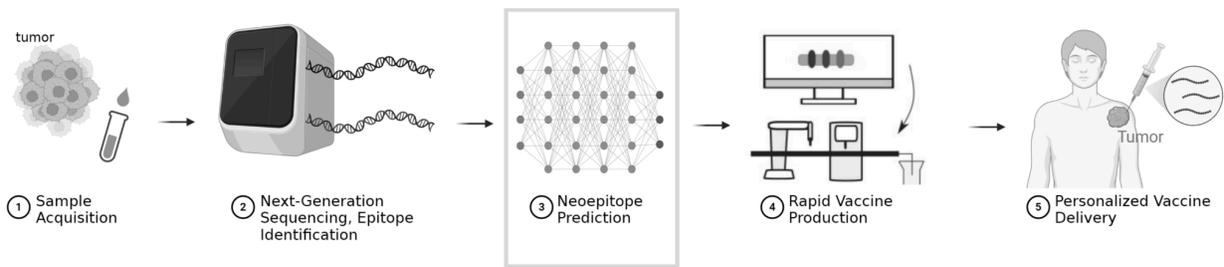

Fig. 1: Personalized Cancer Vaccine Development (created with Biorender)

A small sample of the tumor is taken for next-generation sequencing, the epitopes are then identified through processing including mutational analysis, then computational models are used to predict out of the epitopes present on the tumor, which ones would actually bind to the MHC of the APCs and elicit an immune response. Not all of the epitopes present on a tumor bind to the MHC of the APCs spontaneously due to various chemical properties such as intermolecular forces and orientation, so the prediction of which neoepitopes would bind and successfully elicit an immune response is vital.

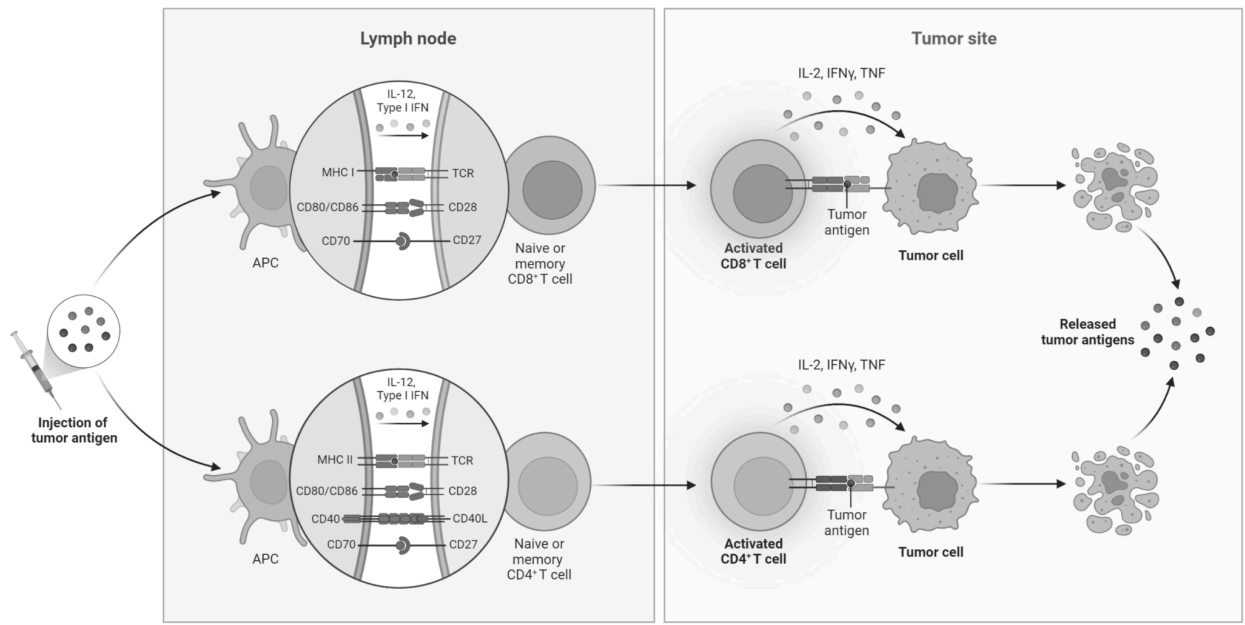

Fig. 2: Biological Mechanism of Vaccine-Induced Immune Response (created with Biorender)

The goal of this project was to create an improved machine learning model for oncologists to use during treatment development towards improved outcomes for cancer patients. Promising results from clinical trials by Moderna and Merck suggest that the future of personalized cancer vaccines is fast approaching, and to assist in the efficient delivery of the vaccines, improved computational models are necessary [4].

## 2.3 Current Models

State-of-the-art models for predicting neoepitope-MHC binding are very large (NetMHCPan-4.0 has 40 Feed Forward Neural Networks, each with 60-70 hidden layers) and require high computational power and time. They are also not commonly used in treatment due to their low relative accuracy on new testing data (NetMHCpan 4.0 - 90%, HLAthena - 85%, MHCFlurry - 81%) [13,16,17].

The majority of these models implement a feed-forward neural network architecture because of their proven ability to be better predictors in classification and regression problems than earlier models such as logistic regression or random forest [11]. Feed-Forward Neural Networks are often able to achieve strong predictive results when trained on non-sequential data [14]. Recurrent Neural Networks are a novel neural network framework and are able to capture and utilize memory in order the make more accurate predictions on sequential data. Since the sequence of the peptides is sequence data, it was hypothesized that a recurrent neural network architecture could be useful in making the neoepitope-MHC binding predictions.

Only two models currently exist that use RNNs in their architecture [12, 10]. Though their models had high predictive power, they may not be useful for the development of personalized cancer vaccines because of their lack of generalization between various HLA classes and binding peptide types. For this reason, NEO (Novel Epitope Optimization) was created.

# 3 Materials and Methods

## 3.1 NCI Dataset

The data used in this project was sourced from a clinical trial approved by the institutional review board of the National Cancer Institute [7]. The 70 individuals chosen for the clinical trial had at least one tumor where whole-exome and whole-transcriptome sequencing was performed with at least one HLA class I-restricted epitope. The dataset had 2,433,281 unique epitopes consisting mainly of mutant residue and the 12 amino acids upstream and downstream of the mutation (resulting in sequences of 25 amino acids in length) [7]. The individual neoepitopes of base pair lengths of 8-12 amino acids from the tumors of the patients were also in the dataset. A subset of the 25 peptide sequences was screened and evaluated for immunoreactivity by co-culturing autologous antigen-presenting cells expressing both the 25 and 8-12 peptide sequences [7]. Since the epitope status was listed as either 0.0 or 1.0 (based on whether the epitope had bound to the MHC complex or not), the machine learning model would have to be created for binary classification. Binary classification involves a model that predicts which class, out of two classes, a sample best fits into based on the input features.

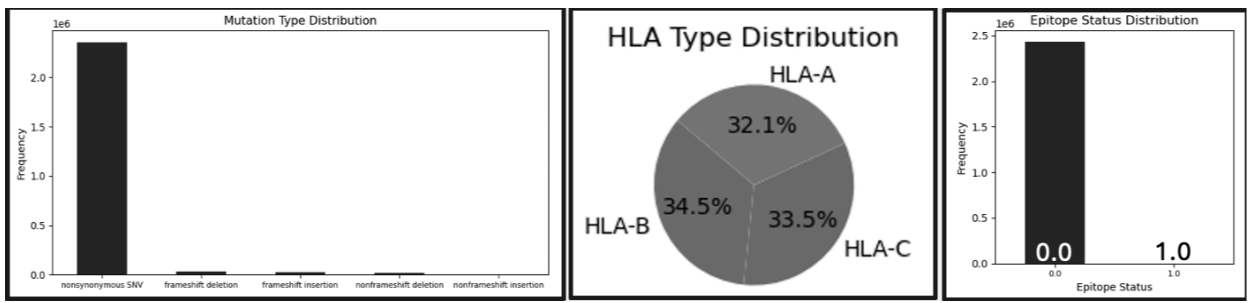

Fig. 3: NCI Dataset Analysis

The data was expanded upon by Gartner et al. 2021 to acquire additional features and this final dataset was expanded upon to acquire even more features from other models to implement the stacking ensemble learning method [6]. The features were preprocessed through a process that included imputing missing values, dropping irrelevant columns, standardization with sci-kit-learn's Standard Scaler module, and label encoding [18]. All of the features used for this model were converted to floats and regarded as non-sequential.

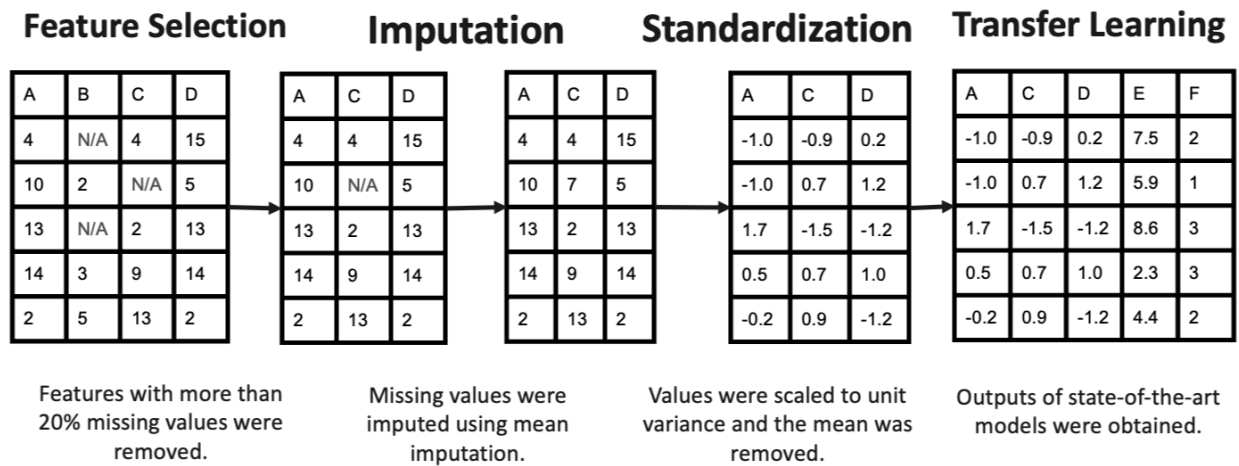

Fig. 4: Data Preprocessing

## 3.2 Model Architecture

### 3.2.1 Feed Forward Neural Network

NEO's architecture was a two-branched, ensemble model with two neural networks. A feed-forward neural network was utilized to learn from patterns from the non-sequential data such as mutation present in RNA sequence, gene expression, and HLA type.

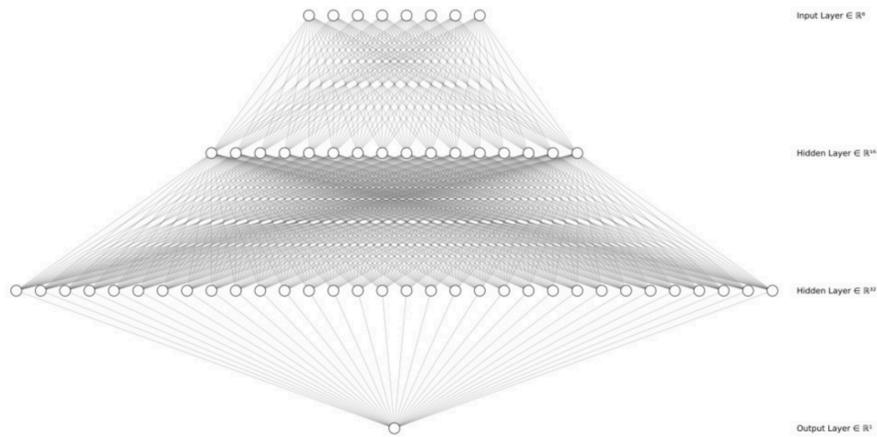

Fig. 5: Architecture of Feed Forward Neural Network with layer sizes of 8, 16, 32, and 1 (created with NN-SVG)

After hyperparameter tuning, the architecture for the feed-forward neural network was 8 neurons in the input layer (for each of the 8 features), 16 neurons in the first hidden layer, 32 neurons in the second hidden layer, and the final output neuron. The optimization process included testing different hyperparameter values to achieve the best performance. Before outputting the final prediction, the value was put into a sigmoid function to output a value between 0 and 1, and a classification threshold of 0.5 was used to determine if the neoepitope was predicted to bind or not. If the value was between 0.5-1.0, the prediction was rounded to 1.0 (positive for binding) but if the value was between 0.0 and 0.5 the value was rounded to 0.0 (negative for binding).

| FFNN Arch. | RNN Arch. | Acc. | AUC Score | Recall |
|---|---|---|---|---|
| 8:16:32:1 | 35:32:32:1 | 99.94% | 0.9166 | 0.8333 |
| 8:16:32:1 | 35:16:1 | 99.91% | 0.6816 | 0.3636 |
| 8:16:32:1 | 35:32:1 | 99.90% | 0.6360 | 0.2727 |
| 8:16:32:32:1 | 35:32:32:1 | 98.16% | 0.9090 | 0.2727 |

Table 1: Hyperparameter Tuning

### 3.2.2 Recurrent Neural Network

An additional RNN model with an LSTM architecture was created and trained on wild-type and mutant peptide sequences in addition to the previous features. Through temporal segmentation, the sequences were broken up into individual amino acids so that they could be used as sequential data. The LSTM architecture was utilized to account for the memory aspect of the sequential peptide sequences [22]. The recurrent neural network had an input layer with all 35 features (the peptide sequences + the non-sequential features + HLA type), 2 Long-Short Term Memory (LSTM) layers with 32 neurons each, and an output layer with one neuron. The output of the model was again inputted into a sigmoid function and a threshold of 0.5 was used for the binary classification. Once again, an output of 1.0 represented binding and 0.0 represented no binding. Both models had a learning rate of 0.001.

To discuss how an RNN is able to utilize memory to learn from the sequential data, the structure of the LSTM layers must be analyzed. In an LSTM, the cell state (C) is what moves the information along. The module can add or remove information from the cell state, as regulated by the gates. In a standard LSTM, there are three gates. The first gate ($f_t$), is the forget gate. This gate determines how much of the previous data the neuron wants to forget. The input gate ($i_t$) records how much of the new input the neuron wants to replace the old information. The final gate is the output gate ($o_t$), which is where the actual adding and removing of information occurs. In this way of regulating how much to remember from previous inputs, the LSTM is able to retain sequential information and make informed predictions on new inputs.

The optimizer used was a combination of Adam (Adaptive Moment Estimation) and SGD (Stochastic Gradient Descent) [19]. A batch size of 100 was used to train the model due to the large dataset size. The loss function utilized was Binary Cross Entropy.

## 3.3 SMOTE

A big challenge with building an accurate machine learning model for this data set was the data imbalance. This dataset has only 139 positive samples but over 2 million negative samples for binding. For this reason, the original training data set was unable to provide the model with sufficient information to effectively predict the neoepitope-MHC binding.

To make up for the lack of positive samples in the dataset, the SMOTE technique was used. SMOTE, or Synthetic Minority Over-sampling Technique, is a technique used when there is a class imbalance problem with the training dataset of a machine learning model. The goal of SMOTE is to balance the class distribution so that the model is better able to learn the patterns of the minority class. SMOTE works by first identifying the minority class. Then, it randomly picks a sample from the minority class and identifies the k-nearest neighbors of the selected instance. Once the neighbors are identified, SMOTE creates synthetic samples by interpolating between the selected sample and its neighbors, a process that includes creating a random instance on the line segment between the two neighbors. The process is repeated until the desired balance between the two classes is achieved.

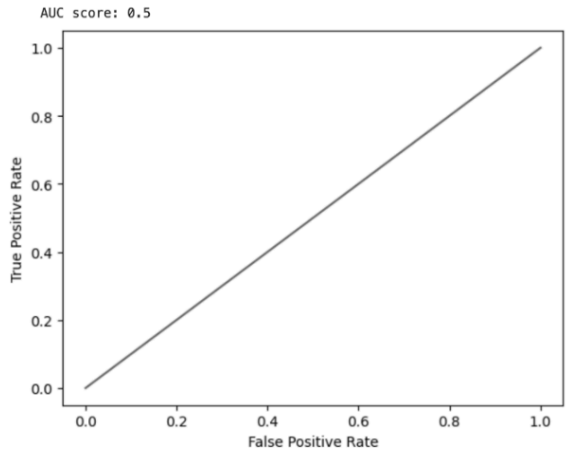

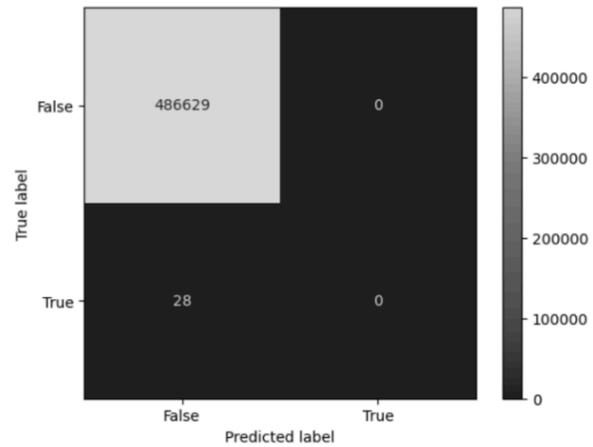

(A)                                                        (B)

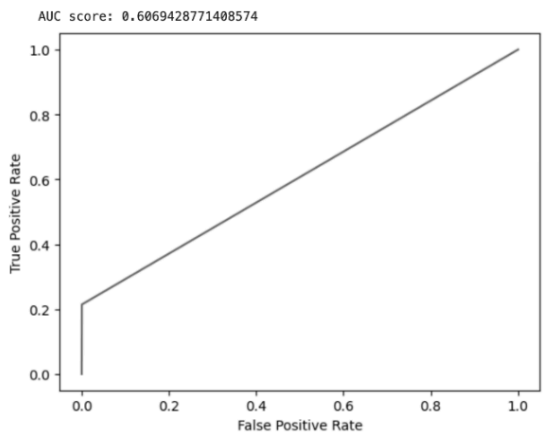

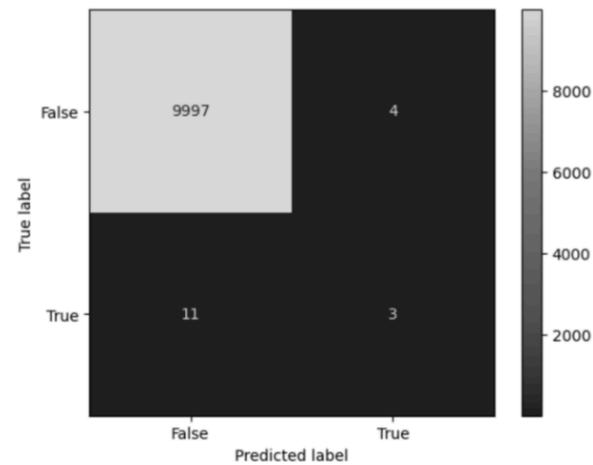

(C)                                                        (D)

Figure 6: a) The ROC curve shows the performance of FFNN prior to the addition of SMOTE. Because the model put every sample into the negative category, the model has the same AUC value of 0.5 as a random classified,  b) The confusion matrix of FFNN prior to SMOTE shows that every negative sample was correctly predicted as negative but every positive sample was also predicted as negative, resulting in an accuracy of 99.99%, c) The ROC curve shows the performance of the RNN without SMOTE. The model is very similar to the FFNN and performed nearly like a random classifier, d) The confusion matrix for the RNN shows that the model correctly predicted 3 of the positive samples as positive and mislabeled 11 positive samples as negative and 4 negative samples as positive.

SMOTE was run on only the training samples to avoid testing on synthetic samples. The ratio between the positive and negative samples in the training set was brought to 1:1.

## 3.4 Model Aggregation

The aggregation technique for merging the results of the two models together was an ensemble with fixed weights (results were averaged together and the threshold of 0.5 was used to determine which classification the ensemble model would choose).

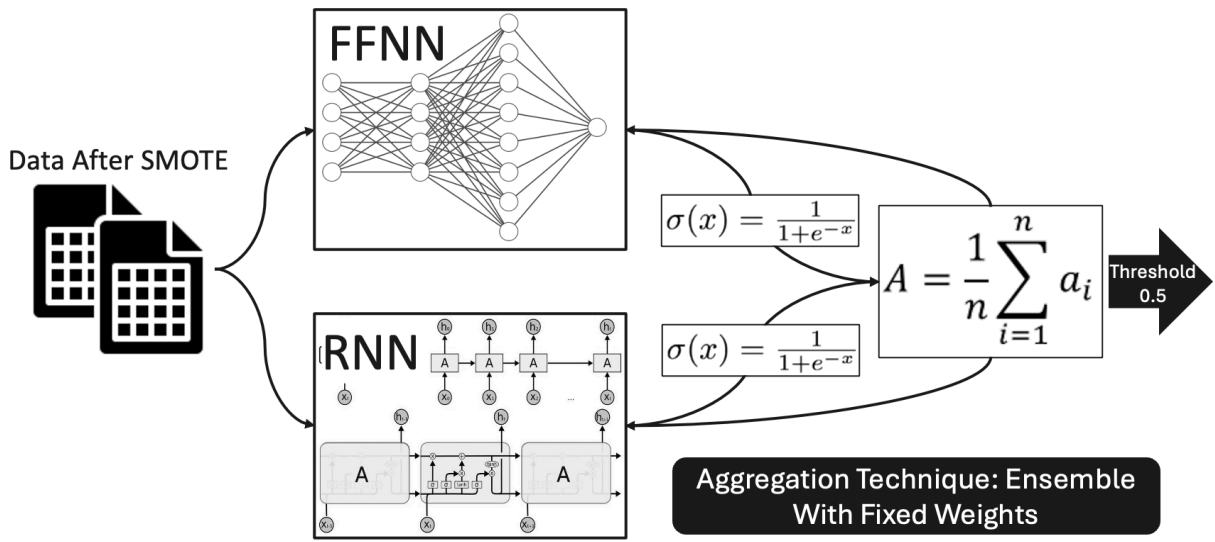

Fig. 7: Complete Model Architecture and Data Flow

# 4 Results

## 4.1 Training Features

The features to be used were determined by running Layer-wise Relevance Propagation (LRP). LRP is a model-agnostic feature importance test that calculates the importance of each feature on the output of a layer. The outputs chosen were those of the FFNN between layer 1 and layer 2.

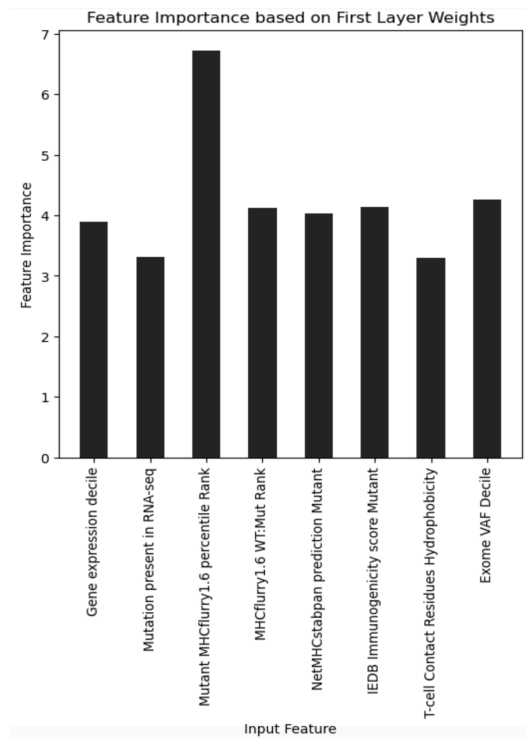

Fig. 8: LRP Results. As shown, all of the features had high importance

To determine the correlation between each of the features, correlation matrices were created. This is done to avoid redundancy introduced if features are highly correlated. Since there was not a strong correlation (>0.75) and all of the features are important, all eight features were used in the final training.

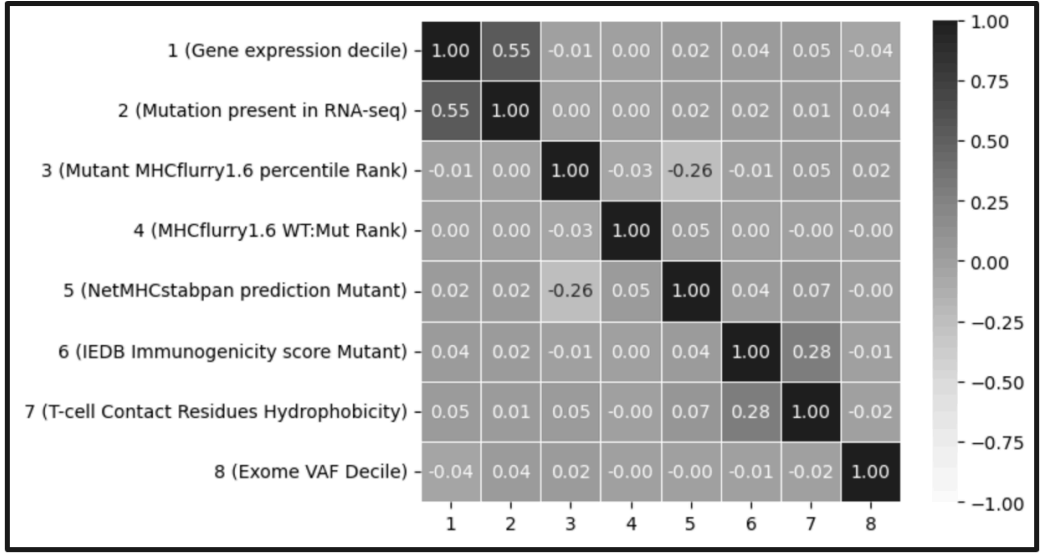

Fig. 9: Correlation Matrix

## 4.2 Testing Results

The final models were trained on ~480,000 samples after synthesizing positive samples with SMOTE. The FFNN was trained for 100 epochs and the RNN was trained for 5 epochs before aggregating the results. The model achieved a loss of 0.0017, an accuracy of 99.98%, a sensitivity of 0.833, a specificity of 0.9999, a precision of 76.92, an F1 score of 0.8, and an AUC score of 0.9166.

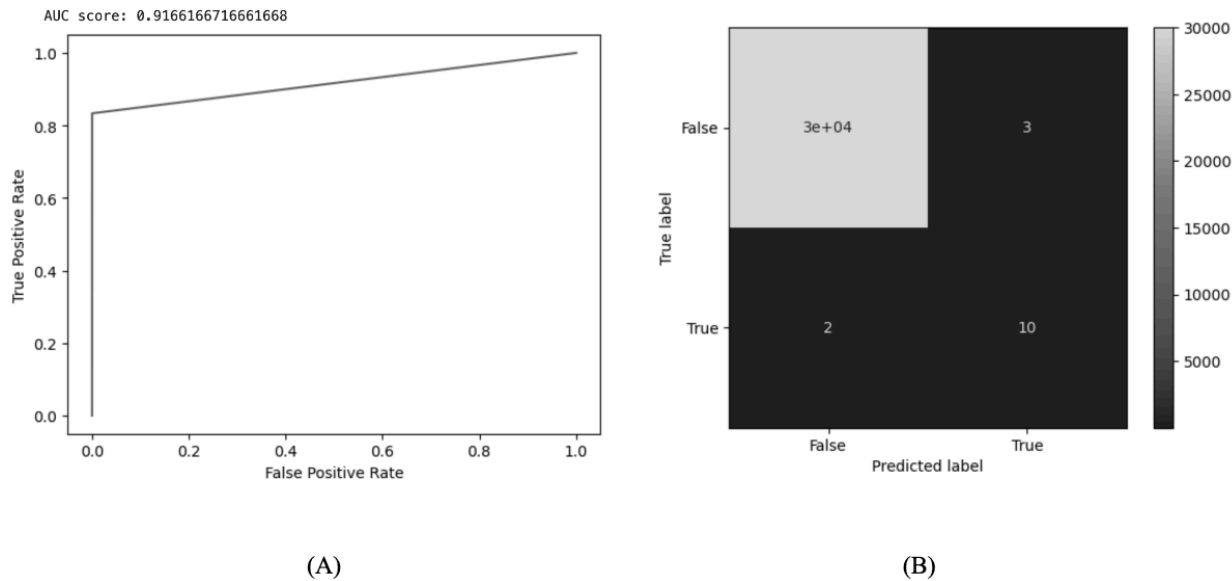

<div align="center">(A)                     (B)</div>

Figure 10: a) The ROC curve shows very accurate predictions with the ensemble FFNN and RNN model with an AUC score of 0.9166. b) The confusion matrix shows that 99.99% (only missed 3 samples out of 30,000) of the negative samples were correctly identified and 83.33% (10/12) of the positive samples were correctly identified.

The ROC curve shows how the true and false positive rates changed at various threshold values. Through this plot, the model's ability to distinguish between positive and negative values is exemplified (AUC = 0.9166). The confusion matrix further validates this.

Though this configuration of parameters was the highest-scoring model across most of the metrics, the highest AUC score from all of the hyperparameter settings was 94.86 and the highest recall was 91.67%. These results indicate that NEO was successfully able to learn from the sequential data provided by the amino acid sequences.

## 4.3 Performance Benchmarking

Through brief comparative metric analysis of the AUC scores (the ability of each model to correctly distinguish between positive and negative samples), NEO outperformed every state-of-the-art model.

| Model: | AUC Score | Model | AUC Score |
|---|---|---|---|
| NEO | 0.917 | NetMHCpan 4.0 BA | 0.749 |
| NMER | 0.911 | MixMHCpred2.02 | 0.730 |
| MHCflurry1.6 | 0.759 | HLAthena | 0.735 |
| NetMHCPan4.0 EL | 0.781 | Random | 0.500 |

Table 2: Comparative Metric Analysis

Additionally, the model was remarkably more efficient. Rather than taking days or weeks to test on samples, NEO only took 69.89 milliseconds to test 40,000 neoepitopes.

| Training set = 640,000 Validation set = 40,000 | Training & Validation of FFNN | Training & Validation of RNN | Testing on 40,000 neoepitopes |
|---|---|---|---|
| Time Taken | 5 minutes & 2 seconds | 16 minutes & 10 seconds | 69.89 milliseconds! |

Table 3: Time Benchmarking

# 5 Discussions and Conclusion

Overall, NEO can revolutionize personalized vaccine production by accurately predicting which neoepitopes should be used in the vaccines. NEO provides hope for the usage of Recurrent Neural Networks when developing predictive models with peptide sequence data, especially in cancer biology. NEO, unlike many state-of-the-art models, doesn't require specific data about the orientation of the neoepitopes. Using NEO, oncologists don't need to wait for days or even hours to get results from a computational model and vaccines can be developed with higher efficiency and efficacy.

To further this research, more relevant features could be extracted from the sequencing data and trained on, such as mutational signatures. Moreover, a larger dataset with more patient variety could be used such as the TESLA dataset [17, 21]. Since this is only the third neoepitope-MHC machine learning model to use RNNs in the prediction model, it shows that using RNNs and peptides as sequential data rather than simply performing label encoding directly on the entire sequence can increase model accuracy and make better predictions.

Additionally, if a system with a larger RAM was provided, even more effective imputers could be used for imputation. Rather than using sci-kit-learn's Standard Imputer, a model with more RAM could impute empty values with a K-Nearest Neighbors imputer, which would identify the nearest neighbors to an empty datapoint to replace the empty value rather than just simply replacing with the mean as Standard Imputer does [18]. Furthermore, access to stronger GPUs and TPUs could improve the ability to test the model with more validation techniques and the implementation of additional libraries. The model could also be expanded upon by utilizing additional machine learning techniques, such as L1 and L2 regularization. A user interface could also be developed for easier usage for oncologists.

This study plans to further validate the model *in vitro* by checking for immunogenic effects through MTT, gene, and apoptosis assays. According to a comparison of *in silico* and *in vitro* techniques for prioritization of tumor-rejecting neoepitopes by Fritah et al., *in vitro* testing results in more efficacious vaccines [5]. Though NEO has already been validated *in silico*, these pre-clinical trials would give doctors the confidence to utilize NEO in the medicinal setting. By harnessing the power of recurrent neural networks and memory as well as the predictive abilities of feed-forward neural networks, neoepitope-MHC binding predictions are now more accurate than ever. With this model, oncologists can find the best neoepitope candidates to include when manufacturing personalized cancer vaccines, suggesting that the future may be bright for personalized cancer immunotherapy and personalized medicine.

# 6 Selected Bibliography